\documentclass{bmvc2k}

\usepackage{makecell}
\usepackage{algorithm}
\usepackage[noend]{algpseudocode}
\usepackage{dsfont}
\usepackage{wrapfig}
\usepackage{amsfonts}
\usepackage{booktabs}

\usepackage{pifont}
\newcommand{\cmark}{\ding{51}}
\newcommand{\xmark}{\ding{55}}

\usepackage[subtle]{savetrees}

\DeclareMathOperator*{\argmax}{\arg\max}

\title{Maskomaly:\\Zero-Shot Mask Anomaly Segmentation}

\addauthor{Jan Ackermann}{ackermannj@ethz.ch}{1}
\addauthor{Christos Sakaridis}{csakarid@vision.ee.ethz.ch}{2}
\addauthor{Fisher Yu}{fisheryu@ethz.ch}{1,2}

\addinstitution{
 Department of Computer Science\\
 ETH Zurich,\\
 Switzerland
}
\addinstitution{
 Computer Vision Lab\\
 ETH Zurich,\\
 Switzerland
}

\runninghead{Ackermann, Sakaridis, Yu}{Maskomaly}

\def\etal{\emph{et al}\bmvaOneDot}

\begin{document}

\maketitle

\vspace{-15pt}
\begin{abstract}
\sloppy{We present a simple and practical framework for anomaly segmentation called Maskomaly.
It builds upon mask-based standard semantic segmentation networks by adding a simple inference-time post-processing step which leverages the raw mask outputs of such networks. Maskomaly does not require additional training and only adds a small computational overhead to inference. Most importantly, it does not require anomalous data at training. We show top results for our method on SMIYC, RoadAnomaly, and StreetHazards. On the most central benchmark, SMIYC\footnote{Public Benchmark at \url{https://segmentmeifyoucan.com/leaderboard}}, Maskomaly outperforms all directly comparable approaches. Further, we introduce a novel metric that benefits the development of robust anomaly segmentation methods and demonstrate its informativeness on RoadAnomaly. 
}
\end{abstract}

\vspace{-15pt}
\section{Introduction}
\label{sec:intro}
\vspace{-5pt}

Anomaly detection is the task of identifying whether one data point belongs to a set of inlier classes that have been seen during the training.
Recently, Fan \etal~\cite{fan2022gout} have demonstrated how difficult anomaly detection is from a theoretical viewpoint. Nevertheless, it is an essential component of many real-world systems operating in safety-critical settings, such as autonomous cars. In order to achieve fully automated driving, autonomous cars need to understand when an anomaly is present and where the anomaly is located in the scene. The latter task is significantly more challenging, as a model needs not only to output one single anomaly score per scene but a dense map of pixel-level scores. With more diverse datasets~\cite{chan2021segmentmeifyoucan,blum2021fishyscapes,lis2019detecting,hendrycks2019scaling} and more complex training paradigms~\cite{liang2018enhancing,lis2019detecting,besnier2021triggering,chan2021entropy}, current approaches have already achieved high scores on relevant benchmarks. However, the current state of the art has not yet reached a level of accuracy that would allow deployment in real-world settings. Given these facts, one might expect that an even more complex training pipeline is needed to achieve results of practical utility.

Nonetheless, we propose a very simple yet effective zero-shot method which requires no training on anomalous data. Our method, Maskomaly, builds upon mask-based semantic segmentation networks, such as Mask2Former~\cite{cheng2022masked}, by post-processing their raw mask predictions to compute a dense anomaly heatmap, only adding a small computational overhead at inference. To the best of our knowledge, we are the first, along with the concurrent work in~\cite{grcic2023advantages, nayal2022pixels}, to explore the utility of mask-based semantic segmentation networks for anomaly segmentation. Our key insight is that mask-based networks trained for standard semantic segmentation already learn to assign certain masks to anomalies. Even though such masks are discarded by default when generating semantic predictions, we show that they can be leveraged for inference on images potentially containing anomalies to achieve state-of-the-art results in anomaly segmentation. Although Maskomaly is an intuitive extension of Mask2Former, we show that computing the anomaly heatmap from the mask outputs of the latter is nontrivial and justify our proposed components through proper ablations. On multiple anomaly segmentation benchmarks, Maskomaly beats all state-of-the-art methods that do not train with auxiliary data and most methods that perform such training. We evidence the generality of our key insight and method across different backbones. Finally, we present a new metric that promotes robust anomaly prediction methods tailored for real-world settings.

\begin{figure}
    \centering
    \includegraphics[width=\textwidth]{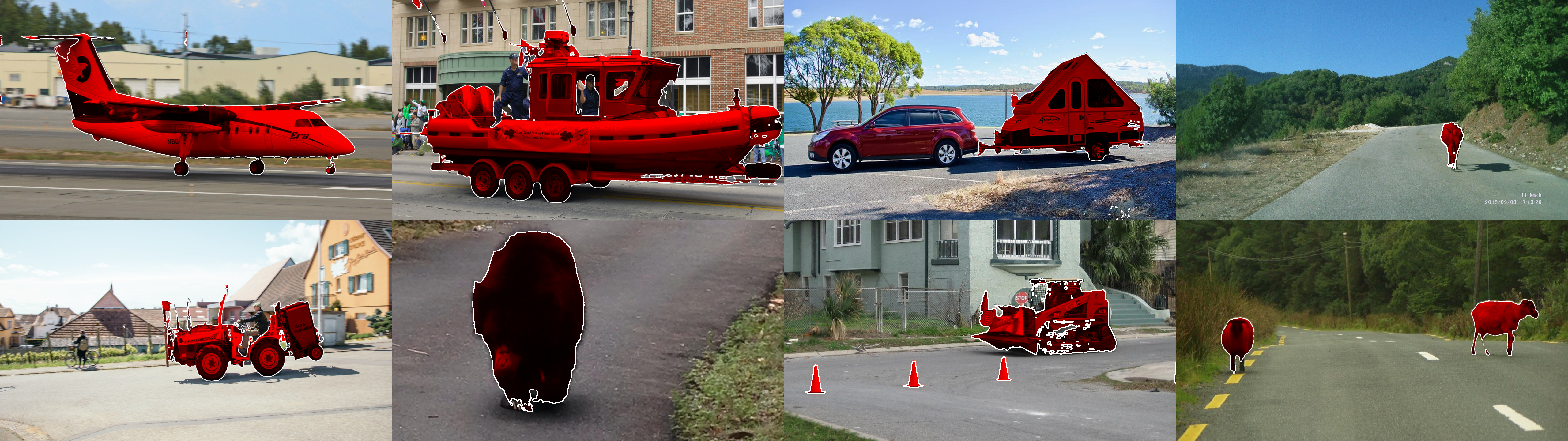}
    \vspace{-15pt}
    \caption{Maskomaly results on the SMIYC~\cite{chan2021segmentmeifyoucan} test set. Segmentations are shown in red and obtained by thresholding at 0.9.}
    \vspace{-15pt}
    \label{fig:teaser}
\end{figure}

\vspace{-10pt}
\section{Related Work}
\label{sec:related}
\vspace{-5pt}

\paragraph{Anomaly Detection:}
Identifying samples that deviate from a known probability distribution is an old problem~\cite{hawkins1980identification}, generally referred to as anomaly detection. Early works identify the uncertainty in the model and utilize ensembling~\cite{lakshminarayanan2017simple}, input perturbations~\cite{liang2018enhancing}, and max-softmax probability~\cite{hendrycks2017baseline}. More encouraging results are achieved by methods that estimate the likelihood with a generative model~\cite{nalisnick2019deep, serra2020input, zhang2021understanding} or that train a model discriminatively with negative data~\cite{liu2020energy, hendrycks2018deep, bevandic2019simultaneous, grcic2021dense, lee2018training}. This is also referred to as training with auxiliary data.

\vspace{-5pt}
\paragraph{Anomaly Segmentation} aims to predict a dense map of outlier probabilities for a given image. Works such as~\cite{oberdiek2020detection, blum2021fishyscapes} provide methods that average over multiple predictions. Another line of work estimates the uncertainty by computing the dissimilarity between the input and a resynthesized image~\cite{di2021pixel, lis2019detecting, vojir2021road}. The synthesized output is conditioned on the semantic segmentation of the input.
A recent work~\cite{besnier2021triggering} trains a network to predict errors of a frozen semantic segmentation network. To stimulate failure, they attack the network with FGSM~\cite{goodfellow2014explaining}. 
GMMSeg~\cite{lianggmmseg} uses GMMs to capture class-conditional densities and train the dense representation in a discriminative manner. 
Methods that train on auxiliary data include Pebal~\cite{tian2022pixel}, which jointly optimizes their novel pixel-wise anomaly abstention learning and energy-based models, DenseHybrid~\cite{grcic2022densehybrid}, which utilizes discriminative and generative modeling simultaneously, and Max.\ Entropy~\cite{chan2021entropy}, which learns to predict high entropy at anomalous regions and uses a meta-classifier to reduce the number of false positives.

\vspace{-5pt}
\paragraph{Mask-Based Approaches:}
The outputs of mask-based segmentation networks have been leveraged in~\cite{heidecker2021towards}, which uses the uncertainty in the instance predictions to detect bounding boxes of anomalies. \cite{breitenstein2021detection} predicts whether an anomaly is present based on the count of instance proposals. Unlike these methods, our method predicts a dense anomaly map for an image, assigning a continuous score to each pixel.
Moreover, \cite{franchi2020one} segments anomalies by rejecting regions with known objects. Distinctly, we do not train one binary classifier per semantic class but use one semantic segmentation model. Further, rejecting known regions is only one part of our method, as we also utilize ``anomalous'' masks as positive predictions.
A concurrent work~\cite{grcic2023advantages} considers Mask2Former for anomaly segmentation. They contribute a score combined with a standard detector~\cite{hendrycks2019scaling}. Furthermore, another work~\cite{nayal2022pixels} contributes a new metric for training to detect outliers. By contrast, our work contributes a complete post-processing method, does not use auxiliary data, and views masks as cohesive units.

\begin{figure}
    \centering
    \includegraphics[height=4.5cm]{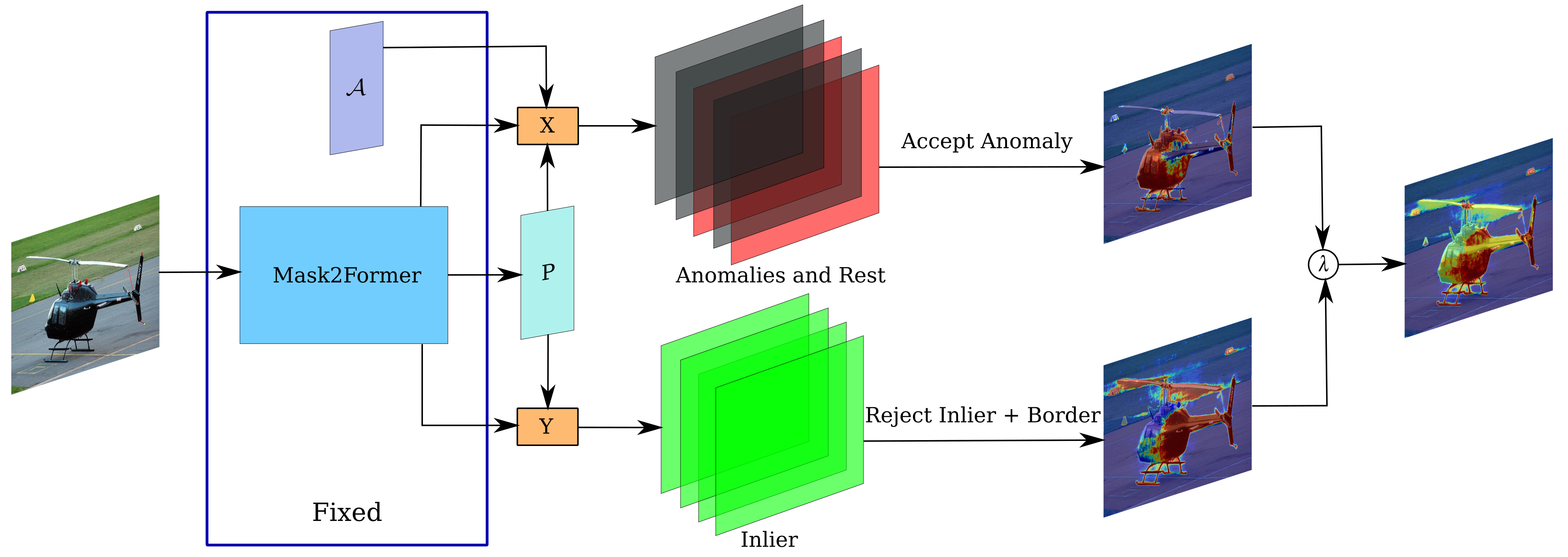}
    \caption{Maskomaly computes the anomaly segmentation by combining two separate predictions. Based on the outputs of class distribution $P$, module Y chooses which masks belong to inliers and constructs a prediction by assigning low values within inlier masks or across their shared borders. X constructs a second prediction by assigning high probability to pixels included in masks which were found to predict anomalies on a validation set. This set of masks is denoted by $\mathcal{A}$. The final prediction is obtained by interpolating the two initial ones with parameter $\lambda$.}
    \label{fig:overview}
    \vspace{-10pt}
\end{figure}

\vspace{-10pt}
\section{Method}
\label{sec:method}
\vspace{-5pt}

Given a set of $C$ semantic inlier classes, our goal is to find an algorithm that receives an RGB image of dimension $H\times{}W\times{}3$ and outputs a dense anomaly segmentation $o = [0,1]^{H\times{}W}$. It should assign high values to anomalous regions and low values to inliers. Anomalies are objects that do not fit into any of the $C$ classes.
For our method, we utilize intermediary outputs of any mask-based segmentation networks. In particular, we work with a set of soft-masks $M = \{m_1,...,m_N\}$ and a set of probability vectors $ P = \left\{p_1,...,p_N \right\}$, where $N$ is the number of masks and $C+1$ is the label of the void class.
A pixel $[i,j]$ belongs to a mask $m_k$ if and only if $k = \argmax_{1\leq n \leq N} m_n[i,j]$. Moreover, a mask $m_k$ belongs to an inlier class if and only if $\argmax_{1\leq l \leq C+1} p_k[l] \neq C+1$.
In the following sections, we use Mask2Former~\cite{cheng2022masked} with a Swin-L backbone and trained on Cityscapes~\cite{cordts2015cityscapes} because it was the best-performing mask-based segmentation network at the time of writing. However, our method is not restricted to a specific segmentation model.
\vspace{-5pt}

\subsection{Masks as Binary Classifiers}
\label{sec:method:rejection}
\vspace{-5pt}

Our first idea is conceptually simple: The segmentation network outputs class probabilities and mask scores for each query. For the $C$ known classes in an image, we expect it to output at least one mask. We confirmed this intuition by evaluating the average IoU of all semantic classes on the test set of Cityscapes~\cite{cordts2015cityscapes}.
Due to the high IoU scores, we interpret each mask as a binary classifier. For a mask $m_i$, this means that there is a set $s_{i}^{T_{\text{mask}}}$ of 2D coordinates that belong to it with probability higher than threshold $T_{\text{mask}}$ and its complement $\overline{s_{i}}^{T_{\text{mask}}}$. Under this assumption, a pixel belongs to an anomaly if and only if it is inside $O = \bigcap_{i\in \mathcal{I}} \overline{s_{i}}^{T_{\text{mask}}}$. This is the intersection of all complements of inlier masks. Here, $\mathcal{I}$ denotes the set of indices of all masks which belong to an inlier class according to $P$. A straightforward way to predict anomalies would be to assign $o[i,j] = \mathds{1}\{[i,j]\in O\}$. To avoid overconfident predictions, we scale the output by the uncertainty of the dominant mask and receive a softer assignment:
\vspace{-2pt}
\begin{equation}
\label{eq:rejection}
    o_{\text{reject}}[i,j] = \min_{n \in \mathcal{I}} (1 - m_n[i,j] \cdot \max_{1\leq l \leq C} p_n[l]).
\end{equation}

We observe that other methods assign high outlier scores at the boundaries of two inlier classes. Working with masks, we find these areas by computing the pairwise intersection of inlier class masks after thresholding them at $T_{\text{b}}$ and assign these areas an outlier score $\epsilon_{\text{b}}$.

\vspace{-5pt}
\subsection{Anomalous Masks}
\label{sec:method:anomalous}
\vspace{-5pt}

Despite the simplicity and theoretical effectiveness of the above rejection-based approach, we must account for possible failure cases, such as large domain shifts. Similar to~\cite{grcic2023advantages}, we observed that on FishyScapes LaF~\cite{blum2021fishyscapes}, over $50\%$ of the outlier pixels belong to an inlier mask. Although the analysis also shows that inlier pixels belong to inlier masks with over $90\%$, these findings suggest the sub-optimal performance of $ o_{\text{reject}}$.

\begin{algorithm}
\caption{Maskomaly,\ Hyperparameters $=\{T_{\text{mask}}, T_{\text{b}}, \epsilon_{\text{b}}, \lambda\}$}\label{alg:base}
\begin{algorithmic}[1]
\Function{Maskomaly}{$M$, $P$, $\mathcal{A}$, $\mathcal{I}$}

\For{$1 \leq n \leq N$}  \Comment{Add inlier masks to $\mathcal{I}$} \label{alg:base:init}
    \If{$(\argmax_{1 \leq l \leq C + 1} p_n[l]) \neq C + 1 \land (\max_{1 \leq l \leq C + 1} p_n[l]) \geq T_{\text{mask}}$}
        \State $\mathcal{I} \gets \mathcal{I} \cup \{n\}$
    \EndIf
\EndFor

\label{alg:base:rej}
\For{$i,j \in H\times{}W$} \Comment{Reject inlier pixels}
    \State $o_{\text{reject}}[i,j]\gets \min_{n \in \mathcal{I}} (1 - m_n[i,j] \cdot \max_{1\leq l \leq C} p_n[l])$
\EndFor

\label{alg:base:bord}
\For{$\{k,n\} \subseteq \mathcal{I}$} \Comment{Reject borders}
    \State $b \gets (m_k > T_{\text{b}}) \odot (m_n > T_{\text{b}})$
    \State $b \gets \min(1 - b + \epsilon_{\text{b}}, 1)$
    \State $o_{\text{reject}} = \min(o_{\text{reject}}, b)$
\EndFor
\label{alg:base:acc}
\For{$i,j \in H\times{}W$} \Comment{Accept anomalous predictions}
    \State $o_{\text{accept}}[i,j]\gets \max_{n \in \mathcal{A}} (m_n[i,j] \cdot p_{n}[C+1])$
\EndFor

\State \Return $\lambda \cdot o_{\text{reject}} + (1-\lambda) \cdot o_{\text{accept}}$ \Comment{Interpolate reject and accept scores}
\EndFunction
\end{algorithmic}
\label{alg:maskomaly}
\end{algorithm}

We go one step further and analyze the behavior and impact of each mask individually. As mask-based segmentation frameworks include mask-loss~\cite{cheng2021per, cheng2022masked}, it is natural to expect that some queries specialize in predicting a single semantic class. We call a query specialized for class $c$ if it predicts a single class $c$ with probability $1-\epsilon$ in over $T_{\text{query}}$ of the cases. We verified this expectation by evaluating the class probabilities on a test set of Cityscapes~\cite{cordts2015cityscapes} with $T_{\text{query}} = 0.9$ and $\epsilon=0.1$ and found queries that predict classes. This gives rise to whether there are also queries that specialize in detecting void-class objects or anomalies.

The main challenge of finding void class queries is differentiating between queries that predict coherent masks of void class objects and queries that predict noise. We found that we cannot distinguish them reliably based on their class probabilities. Instead, we compute the average IoU of each mask with the anomalies on a validation set and extract all queries that generate masks with an average IoU larger than a threshold $T_{\text{IoU}}$. In Section~\ref{sec:backbone}, we show the quantitative results of this method. With this set of masks $\mathcal{A}$, we compute a different prediction that accepts predictions from masks that are associated with anomalies:

\begin{equation}
    o_{\text{accept}}[i,j] = \max_{n \in \mathcal{A}} (m_n[i,j] \cdot p_{n}[C+1]).
\end{equation}

Finally, we present Maskomaly in Alg.~\ref{alg:maskomaly}, combining the modules from Sec.~\ref{sec:method:rejection} and \ref{sec:method:anomalous}.

\vspace{-10pt}
\section{Maximal Detection Margin}
\begin{figure}
    \centering
    \includegraphics[width=\textwidth]{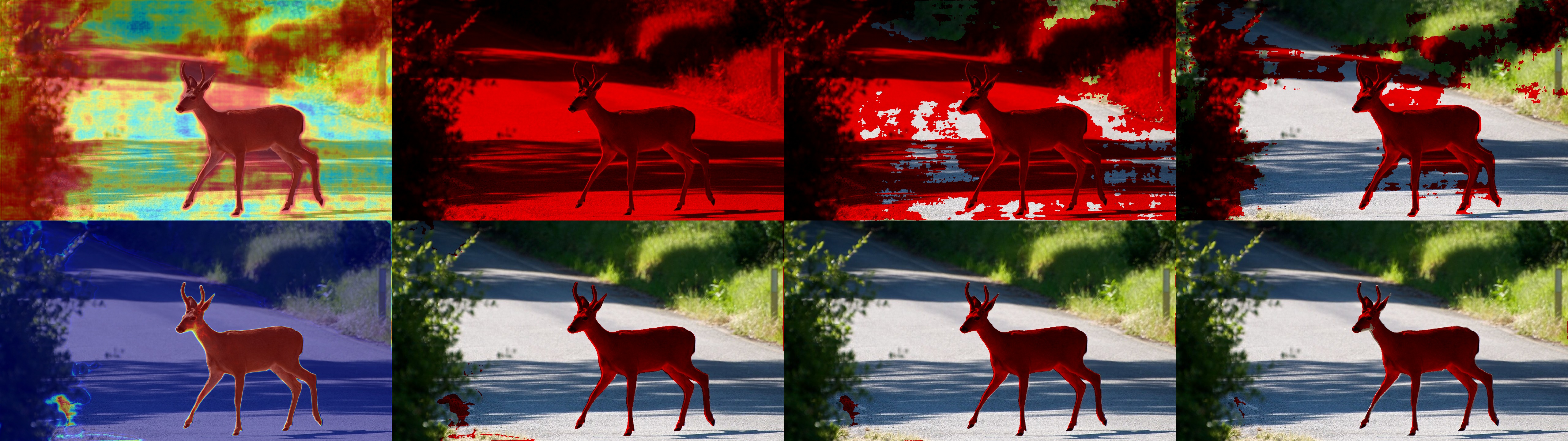}
    \vspace{-15pt}
    \caption{Comparison of Max.\ Entropy~\cite{chan2021entropy} (top) with Maskomaly (bottom) on an image from SMIYC. The first column shows dense predictions; the others show predictions thresholded at 0.3, 0.5, and 0.8. ME achieves 95.5\% AP and Maskomaly 98.9\%.}
    \label{fig:ap_problem}
    \vspace{-25pt}
\end{figure}

\vspace{-5pt}
Chan \etal~\cite{chan2021segmentmeifyoucan} argue that for practical reasons, we need to consider more metrics than the standard Average Precision, FPR95, and AuROC.  Fig.~\ref{fig:ap_problem} illustrates the issue with Average Precision. Despite Max.\ Entropy achieving a high AP of $95.5\%$, it assigns high scores to large parts of the image that are not anomalous. Although it is helpful that a metric is independent of a threshold, it is not applicable in practice. In practice, we need to fix a threshold, and with high probability, it should separate the data well.
Our last contribution is a new metric for anomaly detection that encourages more robust methods.

We present Maximal Detection Margin (MDM) that requires a threshold $T_{\text{margin}}$ and a metric $d$ which measures the difference between a binary ground-truth map $B$ and a soft prediction $A$ thresholded at multiple levels. MDM is computed as the length of the largest interval in which all thresholds achieve at least $T_{\text{margin}}$ regarding metric $d$. Further, it is normalized to $[0,1]$ and computed by:

\begin{equation}
   \text{MDM}_{d}^{T_{\text{margin}}}(A,B) = \max_{0 \leq x < y \leq 1} (y-x)\cdot\mathds{1}\{\forall z \in [x,y]: d(A > z, B) > T_{\text{margin}}\}.
\end{equation}

\vspace{-15pt}
\section{Experiments}
\label{sec:experiments}
\vspace{-5pt}
\subsection{Experimental Setup}
\vspace{-5pt}
\paragraph{Datasets:}
\textbf{SegmentMeIfYouCan} (SMIYC)~\cite{chan2021segmentmeifyoucan} is a road-anomaly dataset consisting of $100$ test and $10$ validation images of anomalies in street scenes with pixel-level annotations. The images show diverse perspectives and anomaly types and have been sourced from the internet. The authors withhold the test set; the scores are only accessible by submitting the method to the official benchmark.
\textbf{FishyScapes}~\cite{blum2021fishyscapes} Static is a dataset based on inserting objects from Pascal VOC~\cite{everingham2009pascal} into Cityscapes~\cite{cordts2015cityscapes} images. Its test set includes $1000$ images, and its validation set includes $30$. We only evaluated our model on the validation set.
\textbf{RoadAnomaly}~\cite{lis2019detecting} consists of 60 images, of which 30 are also part of SMIYC's test set. It shows many scenes with anomalies on streets sourced from the internet.
\textbf{StreetHazards}~\cite{hendrycks2019scaling} is a synthetic dataset recorded inside the Carla Simulator~\cite{dosovitskiy2017carla}. It includes 1500 images and is incredibly challenging due to the high diversity of anomaly types, sizes, and locations.

\paragraph{Metrics:} 
We evaluate other approaches and ours on the standard metrics for anomaly segmentation, Average Precision (AP), False Positive Rate at True Positive Rate of 95\% (FPR95), and Area under Receiver Operator Characteristic (AuROC). On SMIYC, we include their three proposed metrics sIoU gt, PPV, and mean F$_1$~\cite{chan2021segmentmeifyoucan}. On RoadAnomaly, we also evaluate with our novel MDM metric.

\vspace{-10pt}
\sloppy{\paragraph{Hyperparameters:}
For our experiments, we use the semantic segmentation Mask2Former model~\cite{cheng2022masked} with a Swin-L~\cite{liu2021swin} backbone which was trained on CityScapes~\cite{cordts2015cityscapes}. Cheng \etal describe their complete training setup in their work~\cite{cheng2022masked}. For all experiments we use the same SMIYC~\cite{chan2021segmentmeifyoucan} validation set.

Our method is not sensitive to the specific values of its hyperparameters. This is mainly because Mask2Former makes robust predictions. Experiments show that on RoadAnomaly~\cite{lis2019detecting}, over 85\% of the masks are assigned to a class with a confidence of $\geq 0.9$. Thus, $T_{\text{mask}} \leq 0.9$ is a good choice. $93\%$ of the pixels in the entire set of soft mask membership maps for the aforementioned masks have a score of $\leq 0.1$, and from the remaining pixels, $80\%$ have a score of $\geq 0.9$. Thus, the ``border'' pixels can be separated well from the ``core'' mask pixels based on their mask membership scores, and any $T_{b} \geq 0.1$ is a good choice for doing so.
W.r.t.\ $\lambda$, if we assume ``accept'' and ``reject'' are two per-pixel anomaly classifiers that correctly classify pixels with $p > 0.5$, then $\lambda$ should be chosen based on the most likely prior (non-anomalies) due to symmetry. Hence, \emph{any} $\lambda > 0.5$ is a good choice to optimize for AP.

In our experiments we set $\lambda = 0.6$, $T_{\text{IoU}} = 0.25$, $T_{\text{mask}} = 0.3$, $T_{\text{b}} = 0.1$, $\epsilon_{\text{b}} = 0.001$, and $T_{\text{query}} = 0.9$.

We initialize $\mathcal{I}$ to include queries influenced by ambiguous labeling of ground~\cite{cordts2015cityscapes}. We chose the indices based on IoU with street predictions on the SMIYC~\cite{chan2021segmentmeifyoucan} validation set.

\vspace{-5pt}
\subsection{Comparison with the State of the Art}
\label{sec:exp:sota}
\vspace{-5pt}
We include baselines and state-of-the-art methods with segmentation modules and  backbones of similar predictive power in our evaluation. Most approaches, including DenseHybrid~\cite{grcic2022densehybrid}, ObsNet~\cite{besnier2021triggering}, and Max.\ Entropy use DeepLabV3+~\cite{chen2018encoder} with a WideResNet38~\cite{zhu2019improving} backbone. EAM~\cite{grcic2023advantages} uses Mask2Former~\cite{cheng2022masked}. All of these frameworks achieve an IoU of around $83\%$ on Cityscapes~\cite{cheng2022masked, zhu2019improving, chen2018encoder}.

\begin{table}
\centering
\small
\caption{Benchmark results on SMIYC~\cite{chan2021segmentmeifyoucan}. The best method is marked in bold, and the second best is underlined. We separate methods that use auxiliary data during training.}
\vskip 0.1in
\begin{tabular}{llcccccc}\toprule
& & & \multicolumn{5}{c}{Anomaly Track} \\
\cmidrule(lr){4-8}
Method & \makecell{ Segmentation \\ Framework } & \makecell{ Aux. \\ data }  & AP$\uparrow$ & FPR95$\downarrow$ & sIoU gt$\uparrow$ & PPV$\uparrow$ & mean F$_1\uparrow$\\
\midrule
PEBAL~\cite{tian2022pixel} & DLV3+~\cite{chen2018encoder}& \cmark & 49.1 & 40.8 & 38.8 & 27.2 & 14.5\\
SynBoost~\cite{di2021pixel} & VPLR~\cite{zhu2019improving} & \cmark & 56.4 & 61.9 & 34.67 & 17.8 & 10.0\\
DenseHybrid~\cite{grcic2022densehybrid} & DLV3+~\cite{chen2018encoder} & \cmark & 78.0 & 9.8 & 54.2 & 24.1 & 31.1\\
Max.\ Entropy~\cite{chan2021entropy} & DLV3+~\cite{chen2018encoder} & \cmark & 85.5 & 15.0 & 49.2 & 39.5 & 28.7 \\
EAM~\cite{grcic2023advantages} & M2F~\cite{cheng2022masked} & \cmark & 93.8 & \textbf{4.1} & \textbf{67.1} & \textbf{53.8} & \textbf{60.9}\\
RbA~\cite{nayal2022pixels} & M2F~\cite{cheng2022masked} & \cmark & \textbf{94.5} & 4.6 & 64.9 & 47.5 & 51.9\\
\midrule
JSRNet~\cite{vojir2021road}& DLV3~\cite{chen2017rethinking}&\xmark & 33.6 & 43.9 & 20.2 & 29.3 & 13.7\\
DenseHybrid~\cite{grcic2023hybrid} & DLV3+~\cite{chen2018encoder} & \xmark & 51.5 & 33.2 & - & - & - \\
Image Resyn.~\cite{lis2019detecting} & PSP~\cite{zhao2017pyramid} & \xmark & 52.3 & 25.9 & 39.7 & 11.0 & 12.5 \\
ObsNet~\cite{besnier2021triggering} & DLV3+~\cite{chen2018encoder} & \xmark & 75.4 & 26.7 & 44.2 & \textbf{52.6} & \underline{45.1} \\
EAM~\cite{grcic2023advantages} & M2F~\cite{cheng2022masked} & \xmark & 76.3 & 93.9 & - & - & -\\
RbA~\cite{nayal2022pixels} & M2F~\cite{cheng2022masked}  & \xmark & \underline{86.1} & \underline{15.9} & \textbf{56.3} & 41.4 & 42.0\\
Maskomaly (ours) & M2F~\cite{cheng2022masked} & \xmark & \textbf{93.4} &  \textbf{6.9} & \underline{55.4} & \underline{51.5} & \textbf{49.9} \\
\bottomrule
\end{tabular}
\label{tab:smiyc}
\end{table}

\begin{table}
\centering
\small
\caption{Results on RoadAnomaly~\cite{lis2019detecting} and FishyScapes~\cite{blum2021fishyscapes} Static validation set. The best method is marked in bold, and the second best is underlined. We separate methods that use auxiliary data during training.}
\vskip 0.1in
\setlength{\tabcolsep}{3pt}
\begin{tabular}{llccccccc}\toprule
& & & \multicolumn{4}{c}{RoadAnomaly} & \multicolumn{2}{c}{FishyScapes Static} \\
\cmidrule(lr){4-7}\cmidrule(lr){8-9}
Method & \makecell{ Segmentation \\ Framework } & \makecell{ Aux. \\ data }  & AP$\uparrow$ & FPR95$\downarrow$ & MDM$_{\text{F}_1}^{60}\uparrow$ & MDM$_{\text{F}_1}^{70}\uparrow$ & AP$\uparrow$ & FPR95$\downarrow$\\ \midrule
SynBoost~\cite{di2021pixel} & VPLR~\cite{zhu2019improving} & \cmark & 38.2 & 64.8 & 0.0 & 0.0 & 66.4 & 25.6\\
PEBAL~\cite{tian2022pixel} & DLV3+~\cite{chen2018encoder} &\cmark & 45.1 & 44.6 & - & - & \textbf{92.1} & \textbf{1.5}\\
DenseHybrid~\cite{grcic2022densehybrid}& DLV3+~\cite{chen2018encoder} &\cmark & 63.9 & 43.2 & - & - & 60.0 & 4.9\\
M2A~\cite{rai2023unmasking} & M2F~\cite{cheng2022masked} & \cmark & 79.7 & \textbf{13.5} & - & - & - & -\\
Max.\ Entropy~\cite{chan2021entropy} & DLV3+~\cite{chen2018encoder} &\cmark & \textbf{79.7} & 19.3  & \textbf{25.2} & \textbf{9.2} & 76.3 & 7.1 \\
\midrule
ML~\cite{hendrycks2019scaling} & R101~\cite{he2016deep} &\xmark & 19.0 & 70.5  & - & - & 38.6 & 18.3 \\
SML~\cite{jung2021standardized} & DLV3+~\cite{chen2018encoder} &\xmark & 25.8 & 49.7 & - & - & 48.7 & 16.8\\
DenseHybrid~\cite{ grcic2023hybrid}& DLV3+~\cite{chen2018encoder}&\xmark & 35.1 & 43.2 & - & - & 54.7 & 15.5\\
ObsNet~\cite{besnier2021triggering} & DLV3+~\cite{chen2018encoder} &\xmark & 54.7 & 60.0  & 5.1 & 0.0 & 9.4  & 47.7 \\
GMMSeg~\cite{lianggmmseg} & SF~\cite{xie2021segformer} &\xmark & 57.7 & 44.3 & - & - & \underline{82.6} & -\\
EAM~\cite{grcic2023advantages} & M2F~\cite{cheng2022masked}  &\xmark & 66.7 & 13.4 & - & - & \textbf{87.3} & \textbf{2.1}\\
Maskomaly (ours) & M2F~\cite{cheng2022masked} &\xmark & \underline{70.9} & \textbf{11.9} & \textbf{67.1} & \textbf{35.9} & 69.5 & \underline{14.4}  \\
Maskomaly [opt] & M2F~\cite{cheng2022masked}&\xmark & \textbf{80.8} & \underline{12.0} & \underline{62.6} & \underline{20.8} & 68.8 & 15.0  \\
\bottomrule
\end{tabular}
\label{tab:road}
\vspace{-15pt}
\end{table}

Results on SMIYC are shown in Table~\ref{tab:smiyc}. Our method beats all directly comparable approaches in AP, FPR95, sIoU, and mean F$_1$. It is especially noteworthy that our method does not use auxiliary data and improves substantially upon previous such methods as EAM~\cite{grcic2023advantages} and RbA~\cite{nayal2022pixels} which were both designed around Mask2Former~\cite{cheng2022masked} as well. Our performance is even comparable to theirs when trained with anomalous data. Overall, our method ranks third in AP on the entire public SMIYC benchmark. Maskomaly's results on the validation set are visualized and compared to other state-of-the-art methods in Fig.~\ref{fig:smiyc}. Our method can deal with a challenging variety of sizes, shapes, and counts of anomalies. ObsNet~\cite{besnier2021triggering} fails to detect multiple anomalous regions, such as the complete giraffe, the phone booth, and 4 of the birds. Max.\ Entropy~\cite{chan2021entropy} predicts anomalous regions outside the true anomalous objects, such as the people in the tank's background and the street next to the phone booth.

Table~\ref{tab:road} shows our performance compared to state-of-the-art methods on RoadAnomaly~\cite{lis2019detecting} and the validation set of FishyScapes~\cite{blum2021fishyscapes} Static. We observe that on RoadAnomaly, our method outperforms all previous ones that have not trained with auxiliary data by a margin of $4.2$\% on AP and $1.5\%$ FPR95. When considering our method's optimal configuration [opt] as described in Section~\ref{sec:masks}, we even beat Max.\ Entropy by $1.1$\%. On FishyScapes Static, our method ranks second in AP and FPR95 from the methods that do not train with auxiliary data. 
Table \ref{tab:road} also includes the results of four methods on our new metric. Here, we chose F$_1$ because it balances between recall and precision. As it is nontrivial to define a reasonable threshold for F$_1$, we took the scores from Pascal VOC~\cite{everingham2010pascal} and Cityscapes~\cite{cordts2015cityscapes}. We can observe the informativeness of our new metric. Although Max.\ Entropy outperforms our method in terms of AP by $8.8\%$, the unoptimized version of our methods achieves a $41.9\%$ and $26.7\%$ higher score for $\text{MDM}_{\text{F}_1}^{60}$ and $\text{MDM}_{\text{F}_1}^{70}$ respectively.

\begin{figure}
  \centering
  \includegraphics[width=\textwidth]{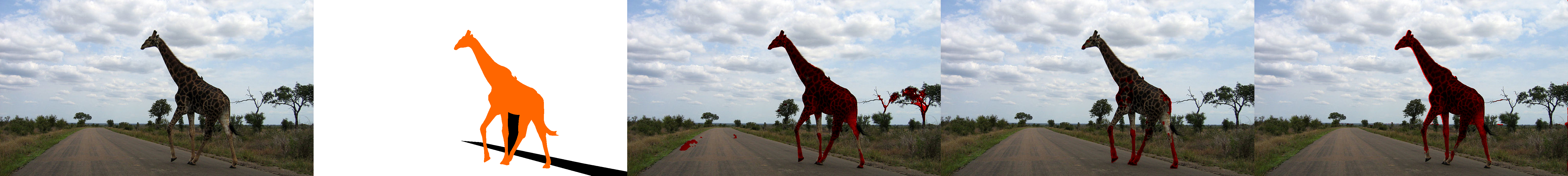}\par
  \includegraphics[width=\textwidth]{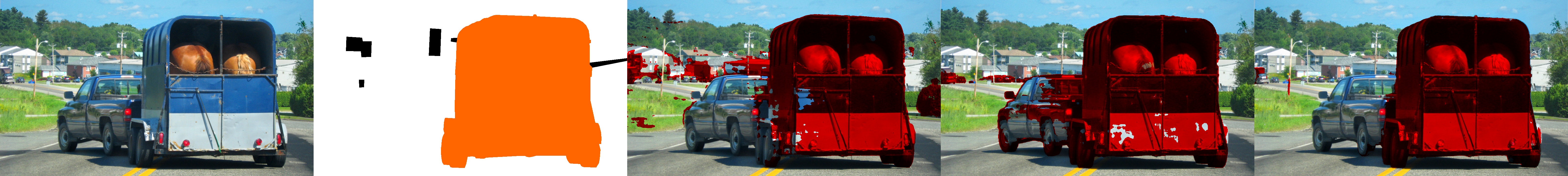}\par
  \includegraphics[width=\textwidth]{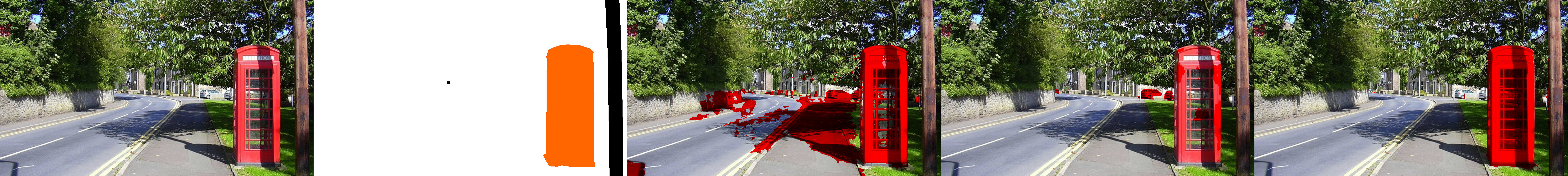}\par
  \includegraphics[width=\textwidth]{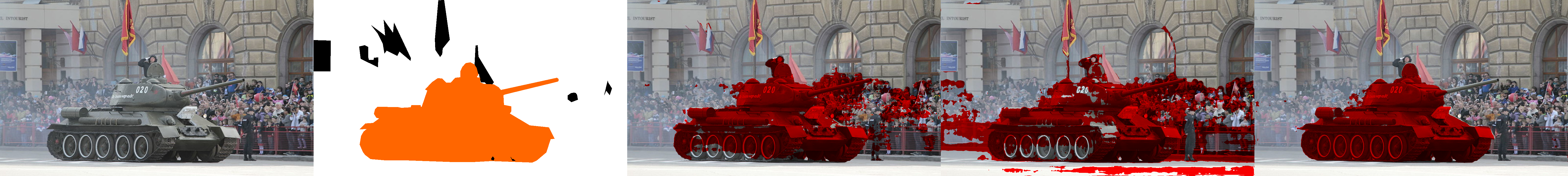}\par
  \includegraphics[width=\textwidth]{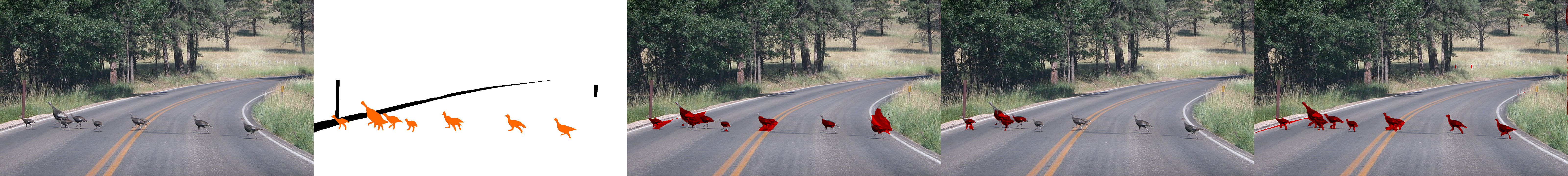}
  \vspace{-15pt}
  \caption{Qualitative results and comparison of our method on SMIYC~\cite{chan2021segmentmeifyoucan} validation set. The first column shows the input. The second shows the ground truth, with the anomalies in orange and void in black. The last three columns show the predicted anomalies of Max.\ Entropy~\cite{chan2021segmentmeifyoucan}, ObsNet~\cite{besnier2021triggering}, and Maskomaly in red. We optimized the threshold to achieve the best F$_1$-score for each image individually and left out void zones.}
  \label{fig:smiyc}
  \vspace{-5pt}
\end{figure}

\begin{table}
\centering
\small
\caption{Results on StreetHazards~\cite{hendrycks2019scaling} test set. The best method is marked in bold, and the second best is underlined. We separate methods that use auxiliary data during training.}
\vskip 0.1in
 \setlength{\tabcolsep}{3pt}
\begin{tabular}{lllccccc}\toprule
& & & & &\multicolumn{3}{c}{StreetHazards (SH)} \\
\cmidrule(lr){6-8}
Method & \makecell{ Segmentation \\ Framework } & Backbone & \makecell{ Aux. \\ data } & \makecell{ SH \\ trained }  & AP$\uparrow$ & FPR95$\downarrow$ & AUC$\uparrow$ \\
\midrule
OH*MSP~\cite{bevandic2021dense} & LDN~\cite{krevso2020efficient} & DenseNet-169 &  \cmark & \cmark & 18.8 & 30.9 & 89.7 \\
Max.\ Entropy~\cite{chan2021entropy} & DeepLabV3+~\cite{chen2018encoder} & WideResNet-38 & \cmark & \xmark & 19.3 & 22.9 & 91.6\\
DenseHybrid~\cite{grcic2022densehybrid} & LDN~\cite{krevso2020efficient} & DenseNet-121 & \cmark & \cmark & \textbf{30.2} & \textbf{13.0} & \textbf{95.6}\\
\midrule
TRADI~\cite{franchi2020tradi} & PSP-Net~\cite{zhao2017pyramid} & ResNet-50 & \xmark & \cmark & 7.2 & 25.3 & 89.2\\
SynthCP~\cite{xia2020synthesize}& PSP-Net~\cite{zhao2017pyramid} & ResNet-101 & \xmark & \cmark & 9.3 & 28.4 & 88.5\\
OVNNI~\cite{franchi2020one} & PSP-Net~\cite{zhao2017pyramid}& ResNet-50 &  \xmark & \cmark & 12.6 & 22.2 & 91.2\\
SO+H~\cite{grcic2020dense} & LDN~\cite{krevso2020efficient} & DenseNet-121 & \xmark & \cmark & 12.7 & 25.2 & 91.7\\
ObsNet~\cite{besnier2021triggering} & DeepLabV3+~\cite{chen2018encoder} & WideResNet-38 & \xmark & \xmark & 12.8 & 25.3 & 90.5\\
DML~\cite{cen2021deep} & PSP-Net~\cite{zhao2017pyramid} & ResNet-50 & \xmark & \cmark & 14.7 & \textbf{17.3} & \underline{93.7}\\
DenseHybrid~\cite{grcic2022densehybrid} & LDN~\cite{krevso2020efficient} & DenseNet-121  & \xmark & \cmark & \underline{19.7} & \underline{17.4} & \textbf{93.9}\\
Maskomaly (ours) & Mask2Former~\cite{cheng2022masked} & Swin-L & \xmark & \xmark & \textbf{23.5} &  19.6 & 91.7\\
\bottomrule
\end{tabular}
\label{tab:sh}
\vspace{-15ptm}
\end{table}

Table \ref{tab:sh} shows the quantitative results of our method compared to state-of-the-art methods on StreetHazards~\cite{hendrycks2019scaling}. Our method beats all methods on AP but the version of DenseHybrid which is trained on StreetHazards and with auxiliary data. We achieve $4.2\%$ higher AP than Max.\ Entropy which was also not trained on the StreetHazards training set. Our method also achieves $3.8\%$ better AP than any method that does not train with auxiliary data even though these methods do not suffer from distribution shift.
\vspace{-15pt}

\paragraph{Inference Speed:}
We measure the computational overhead of our unoptimized implementation on an Intel I7-13700K. For this, we average over 1000 runs of our method for the images from the validation set of SMIYC~\cite{chan2021segmentmeifyoucan}. Our method takes $182$ms on average. We perform the same experiment on an Nvidia RTX4080 with a GPU version that does not perform the border rejection, and it takes $0.48$ms on average.

\vspace{-5pt}
\subsection{Ablations}
\vspace{-5pt}
\paragraph{Individual Components:}
To understand the influence of the individual parts of the algorithm, we remove them one at a time and evaluate the performance of the ablated model. We also included a straightforward baseline that leverages Mask2Former outputs to emphasize more clearly the merit of Maskomaly. This baseline works by assigning 
\vspace{-5pt}
\begin{equation}
    o_{\text{baseline}}[i,j] =
\begin{cases} 
    \displaystyle\max_{1\leq n \leq N} m_n[i,j], & \text{if } \displaystyle\argmax_{1\leq l \leq C+1} p_{\argmax_{1\leq n \leq N} m_n[i,j]}[l] = C+1,  \\
    1-\displaystyle\max_{1\leq n \leq N} m_n[i,j] & \text{otherwise.}
\end{cases}
\end{equation}
\vspace{-5pt}

Intuitively speaking, the baseline checks for each pixel individually whether it belongs to an anomalous mask or not. After that, it assigns the probability of belonging to class $C+1$ or one minus the probability of belonging to its predicted inlier class.

Table \ref{tab:ablating} shows the performance of our ablated models and the baseline (Id 1) on the RoadAnomaly dataset and the test set of SMIYC. We can see that all ablated methods perform better than the baseline. Additionally, each individual component performs better than methods such as PEBAL~\cite{tian2022pixel} or DenseHybrid~\cite{grcic2022densehybrid} without outlier supervision. Rejecting border regions reduces our FPR95 significantly. The largest step is achieved by combining the ideas of rejecting and accepting. This shows that the components complement each other well. Lastly, our mask-based approach allows us to act on masks as a coherent component. We gain significant AP and reduce the FPR95 greatly by initializing $\mathcal{I}$ with masks that we found to predict the ambiguous ground class~\cite{cordts2015cityscapes}, leading us to our complete Maskomaly method (Id 6). The impact of this initialization on RoadAnomaly~\cite{lis2019detecting} is especially high because all anomalies are located on streets or ground.

Fig.~\ref{fig:curves} shows a complete image of our performance.
We can observe that the individual components have a lot less smooth curve than a method such as Max.\ Entropy. This is because we consider masks on a component level and possibly activate entire regions when increasing the threshold slightly. Maskomaly (Id 6) has a smoother curve and its graph lies completely above all individual component combinations (Id 2-5).
Handling borders (Id 4) in addition to only rejecting areas (Id 3), only separate in the higher recall range. In that range, the borderless version loses by a small margin.

\vspace{-5pt}
\paragraph{Method:}
To show the impact of our algorithm decoupled from its backbone we compare it to state-of-the-art methods paired with Mask2Former~\cite{cheng2022masked}. There are three other methods that are build around mask-based segmentation outputs. For a comparison against other methods which were fine-tuned Mask2Former, see the discussion of Nayal et al.~\cite{nayal2022pixels}. \footnote{We did not include the results of RbA~\cite{nayal2022pixels} in the other tables, as the results that RbA~\cite{nayal2022pixels} reports for other methods deviate from what Grcic et al.~\cite{grcic2023advantages} and we report.} In Table \ref{tab:mask_methods}, we show the results of EAM~\cite{grcic2023advantages}, RbA~\cite{nayal2022pixels}, M2A~\cite{rai2023unmasking} and our method on the SMIYC test set. We can clearly see that our method performs by far best when only considering the methods that did not resort to auxiliary data. Even though EAM and RbA gain much by training with anomalous data, our method can keep up in the two most important metrics AP and FPR95 as well as PPV. Especially with regard to AP, our method is only $1.1\%$ behind the current best method.

\begin{table}
\centering
\small
\caption{Results of our ablated models on RoadAnomaly~\cite{lis2019detecting} and SMIYC~\cite{chan2021segmentmeifyoucan} test set. The individual components that are used in each version are ticked. The abbreviations we use are related to particular lines of code in Algorithm~\ref{alg:maskomaly}. Acc.\ refers to $o_{\text{accept}}$  at lines 11--12, Rej.\ refers to $o_{\text{reject}}$ at lines 5--6, Bord.\ refers to lines 7--10, and Init.\ $\mathcal{I}$ refers to lines 2--4. The best method is marked in bold.}
\vskip 0.1in
 \setlength{\tabcolsep}{3pt}
\begin{tabular}{cccccccccc}
\toprule
& & & & &\multicolumn{3}{c}{RoadAnomaly} & \multicolumn{2}{c}{SMIYC} \\
\cmidrule(lr){6-8}\cmidrule{9-10}
Method Id& Acc.\ & Rej.\ & Bord.\ & Init.\ $\mathcal{I}$ & AP$\uparrow$ & FPR95$\downarrow$ & AUC$\uparrow$ & AP$\uparrow$ & FPR95$\downarrow$\\
\midrule
1 & N/A & N/A & N/A & N/A & 18.6 & 54.4 & 74.3 & - & - \\
2 & \cmark & \xmark & \xmark &\xmark & 46.0 & 26.3 & 89.2 & - & - \\
3 & \xmark & \cmark & \xmark & \xmark & 45.3 & 19.3 & 91.3 &  - &  -\\
4 & \xmark & \cmark & \cmark & \xmark & 45.5 & 15.7 & 92.0 &  58.4 & 23.4  \\
5 & \cmark & \cmark & \cmark & \xmark & 63.2 &  14.8 & 94.2 & - & -\\
6 & \cmark & \cmark & \cmark & \cmark & 70.9 &  11.9 & 95.5 & 93.4 & 6.9\\
\bottomrule
\end{tabular}
\label{tab:ablating}
\end{table}

\begin{table}
\centering
\small
\caption{Benchmark results on SMIYC~\cite{chan2021segmentmeifyoucan} of all methods that are designed around mask-based segmentation networks. All methods use Mask2Former~\cite{cheng2022masked} as a backbone. The best method is marked in bold, and the second best is underlined. We separate methods that use auxiliary data during training.}
\vskip 0.1in
\begin{tabular}{lccccccc}\toprule
& & &\multicolumn{5}{c}{Anomaly Track} \\
\cmidrule(lr){4-8}
Method & \makecell{ Aux. \\ data } & retrain  & AP$\uparrow$ & FPR95$\downarrow$ & sIoU gt$\uparrow$ & PPV$\uparrow$ & mean F$_1\uparrow$\\
\midrule
M2A~\cite{rai2023unmasking} & \cmark & \cmark & 88.7 & 14.6 & 55.3 & 51.7 & 47.2\\
EAM~\cite{grcic2023advantages} & \cmark & \xmark & 93.8 & \textbf{4.1} & \textbf{67.1} & \textbf{53.8} & \textbf{60.9}\\
RbA~\cite{nayal2022pixels} & \cmark & \cmark & \textbf{94.5} & 4.6 & 64.9 & 47.5 & 51.9\\
\midrule
EAM~\cite{grcic2023advantages} & \xmark & \xmark & 76.3 & 93.9 & - & - & -\\
RbA~\cite{nayal2022pixels} & \xmark & \cmark & \underline{86.1} & \underline{15.9} & \textbf{56.3} & \underline{41.4} & \underline{42.0}\\
Maskomaly (ours) & \xmark & \xmark &  \textbf{93.4} &  \textbf{6.9} & \underline{55.4} & \textbf{51.5} & \textbf{49.9} \\
\bottomrule
\end{tabular}
\label{tab:mask_methods}
\end{table}

\vspace{-5pt}
\subsection{Hyperparameter Study}
\paragraph{Mask Count:}
\label{sec:masks}
\vspace{-5pt}
In Section~\ref{sec:exp:sota}, we set $T_{\text{IoU}}$ and computed $\mathcal{A}$ in advance for evaluating on the test set to ensure a fair comparison. In this section, we again rank the masks by IoU on the SMIYC validation set and evaluate how the test set scores change when we take the first $n$ masks for $n=1,...,16$. We stop at $16$ because there is no relevant change anymore. Fig.~\ref{fig:hyper} (left) shows that for RoadAnomaly, the optimal configuration for AP would have included only one mask with $80.8\%$ AP, and the optimal configuration for FPR95 would have included 9 with $10.37\%$ FPR95. Nevertheless, it also shows that the method is robust with regard to the number of masks chosen, as the AP does not drop significantly.

\begin{figure}[tb]
    \centering
    \includegraphics[clip,trim=0mm 5mm 0mm 13mm,height=5cm]{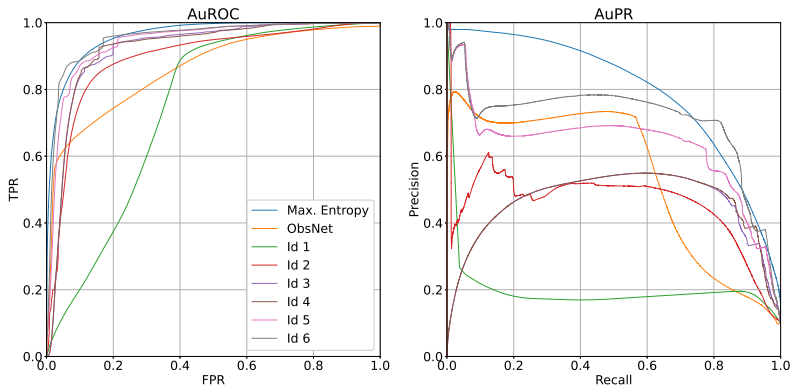}
    \vspace{-5pt}
    \caption{AuROC and AuPR curves of our ablated methods, ObsNet~\cite{besnier2021triggering} and Max.\ Entropy~\cite{chan2021entropy} on RoadAnomaly~\cite{lis2019detecting}. The colors are consistent across the two plots and the Ids are the same as in Table~\ref{tab:ablating}.}
    \label{fig:curves}
\end{figure}

\vspace{-10pt}
\paragraph{Backbone:}
\label{sec:backbone}
Although mask-based segmentation networks such as Mask2Former perform well on inlier classes of Cityscapes~\cite{cheng2022masked}, we need to verify that they generally learn anomaly-predicting masks implicitly.
To show this, we compute the IoU of the masks generated by all available pre-trained semantic segmentation backbones of Mask2Former with anomalies from a control set. We use the SMIYC validation set as the control set as it includes a diverse set of anomalies.
Fig.~\ref{fig:hyper} (right) shows the number of masks that fall into fixed ranges of IoU for all backbones. As in the previous investigation, we cut off our classification after 16 masks. Res101, Swin-S, Swin-B, and Swin-L have masks with a high IoU of more than $40\%$. This provides evidence that a variety of semantic segmentation backbones with large enough capacity implicitly learn masks for anomalies.

\begin{figure}
  \centering
    \includegraphics[height=4.5cm]{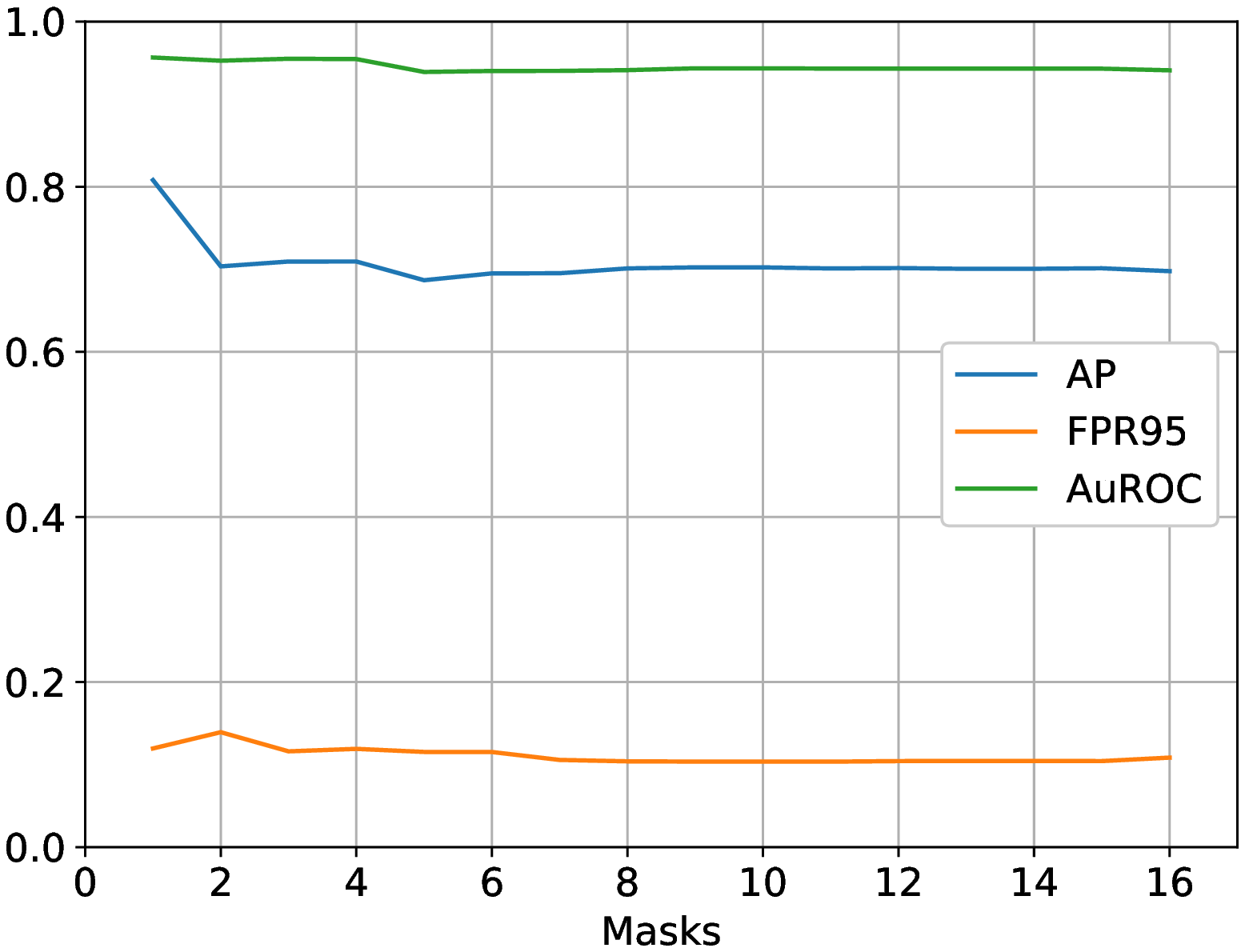}
    \hspace{5pt}
    \includegraphics[height=4.5cm]{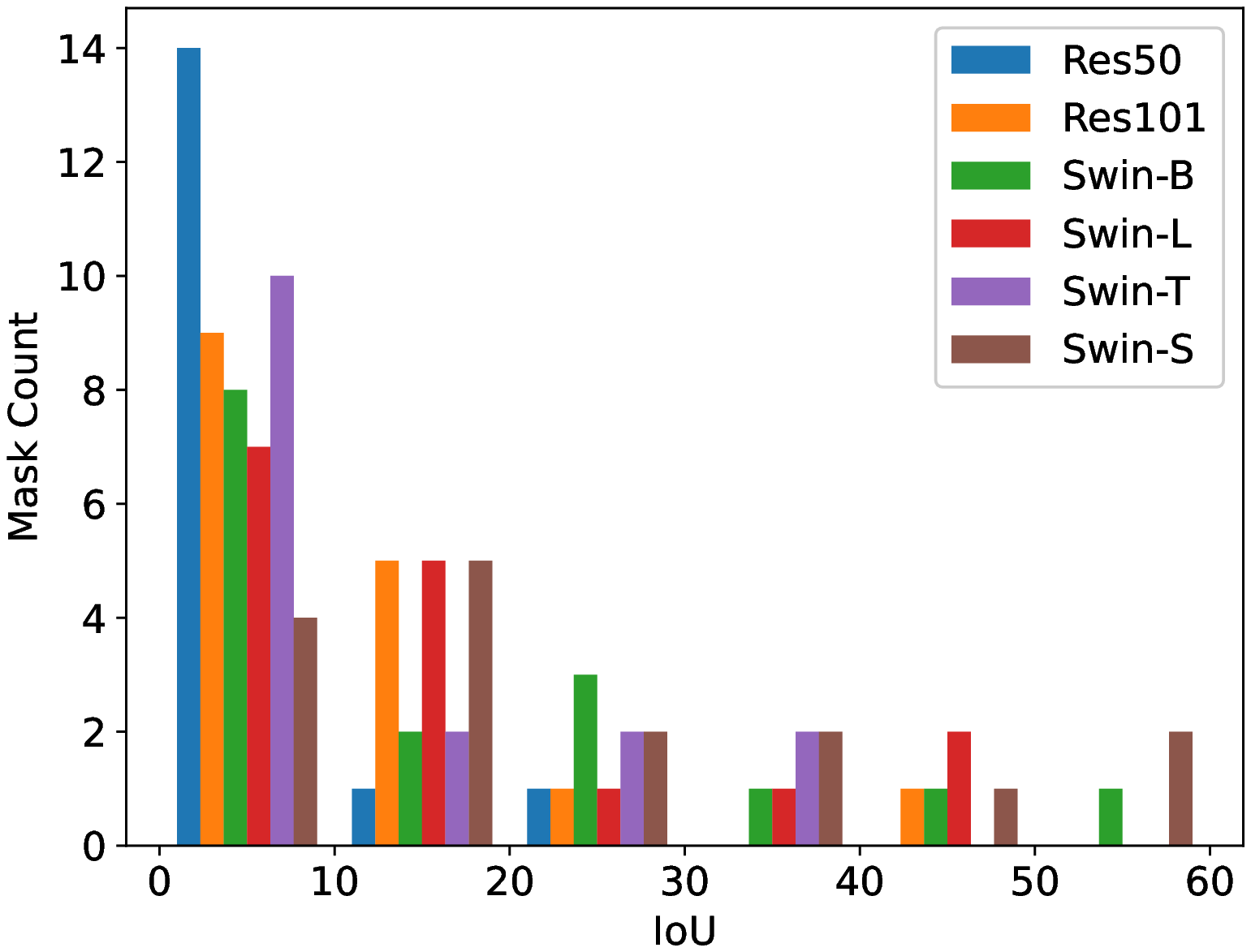}
    \vspace{-5pt}
    \caption{Left: AP, FPR95, and AuROC of Maskomaly for different mask counts on RoadAnomaly~\cite{lis2019detecting}, right: mask counts within different ranges of IoU on SMIYC~\cite{chan2021segmentmeifyoucan} validation.}
    \label{fig:hyper}
\end{figure}

\vspace{-15pt}
\section{Conclusion}
\vspace{-5pt}
\label{sec:conclusion}
We have proposed Maskomaly, a simple yet effective anomaly segmentation method that builds upon pre-trained mask-based semantic segmentation networks. The method is easy to implement, requires no additional training, and sets the new state of the art among methods that do not train with auxiliary data. We have shown the merit of our method against state-of-the-art approaches, ablated its components, and showed the generality of our findings and method across different backbones. 
We have also introduced a novel metric that promotes robust anomaly segmentation methods in view of real-world scenarios. Our results demonstrate the potential of mask-based semantic segmentation networks for anomaly segmentation and provide insights for developing more practical and effective methods for anomaly segmentation in safety-critical systems such as autonomous cars.

\vspace{-10pt}
\bibliography{refs}
\end{document}